\documentclass[twocolumn]{galbot}

\usepackage{microtype}
\usepackage{graphicx}
\usepackage{subcaption}
\usepackage{booktabs}
\usepackage{multirow}
\usepackage{makecell}
\usepackage{pifont}
\usepackage{enumitem}
\usepackage{tabularx}
\usepackage{hyperref}
\usepackage{amsmath}
\usepackage{amssymb}
\usepackage{mathtools}
\usepackage{amsthm}
\usepackage[most]{tcolorbox}
\usepackage[table]{xcolor}
\usepackage{float}

\newcommand{\method}{\textbf{\emph{FutureNav}}}

\definecolor{linecolor1}{RGB}{246, 248, 239}
\definecolor{linecolor2}{RGB}{230, 234, 217}
\definecolor{linecolor3}{RGB}{211, 222, 190}

\title{FutureNav: Unified World-Action Modeling for Vision-and-Language Navigation}

\author[1,2,4,*]{Lingfeng Zhang}
\author[3,4,*]{Zeying Gong}
\author[4,\dagger,\ddagger]{Xiaoshuai Hao}
\author[5]{Haoxiang Fu}
\author[3]{Qiang Zhang}
\author[4]{Mingliang Zhou}
\author[4]{Hangjun Ye}
\author[2]{Xiaojun Liang}
\author[3]{Junwei Liang}
\author[1,\dagger]{Wenbo Ding}
\affiliation[1]{Tsinghua University}
\affiliation[2]{Pengcheng Laboratory}
\affiliation[3]{The Hong Kong University of Science and Technology (Guangzhou)}
\affiliation[4]{Xiaomi EV}
\affiliation[5]{National University of Singapore}
\contribution[*]{Equal contribution}
\contribution[\ddagger]{Project leader}

\contribution[\dagger]{Corresponding authors}
\date{\today}
\page{\url{https://linglingxiansen.github.io/FutureNav/}}

\code{\url{https://github.com/linglingxiansen/FutureNav}}

\correspondence{\url{haoxiaoshuai@xiaomi.com}, \url{ding.wenbo@sz.tsinghua.edu.cn}}

\abstract{
Vision-and-language navigation (VLN) in continuous environments requires an agent to
ground instructions in egocentric observations while maintaining spatial
understanding across long action sequences. Recent navigation foundation models
have shown strong progress by scaling vision-language models (VLMs), but they often
learn navigation primarily as direct action generation, without explicitly
modeling world states or predicting their future evolution. We introduce
\method{}, a VLM-based unified world-action modeling framework for
vision-and-language navigation. Specifically, \method{} jointly encodes text,
visual, and spatial features and feeds them into the LLM, and optimizes four
objectives for simultaneous world and action modeling: an action policy objective
for navigation action prediction, inverse and forward dynamics objectives for
modeling state transitions, and a future generation objective for predicting
future spatial states. This unified architecture strengthens action prediction
while explicitly modeling the world, without sacrificing inference speed.
Extensive experiments show that, with only a 4B-scale backbone, \method{}
achieves state-of-the-art performance on multiple VLN benchmarks and
substantially outperforms prior VLN methods, paving the way toward future
world-action models for VLN. We will release the code and models to support
future research.
}

\begin{document}
\maketitle
\pagestyle{plain}

\section{Introduction}
\label{sec:intro}
Vision-and-language navigation (VLN) requires an embodied agent to follow
natural language instructions and navigate to target locations in unseen 3D environments
\cite{anderson2018vision,krantz2020beyond,ku2020room}. In continuous
environments, this task is especially challenging because the agent must ground
linguistic goals in egocentric observations, retain observation history,
maintain spatial understanding over long action sequences, and execute low-level
actions without accumulating irreversible errors. Early VLN-CE systems therefore
relied on explicit navigation structures, such as waypoint prediction,
topological planning, semantic maps, and learned spatial memories, to connect
language grounding with physically executable actions
\cite{krantz2021waypoint,hong2022bridging,hong2023learning,wang2023gridmm,wang2023dreamwalker,zhang2025mapnav,liu2025toponav,zhang2024trihelper,zhang2024multifloor,gong2025stairway,zhang2025socialnavmap,zhang2026walkwithme,kong2026robosense,zhang2025xiaomievad,xiao2025socialrisk,hao2025mapfusion,hao2026synergistic,hao2026embodied,zhang2026embodiedplan}.
These methods indicate that successful embodied navigation requires more than
recognizing objects or instruction phrases: an agent must also understand the
state of the surrounding world and how that state evolves under its actions.

\begin{figure*}[!ht]
    \centering
    \includegraphics[width=\linewidth]{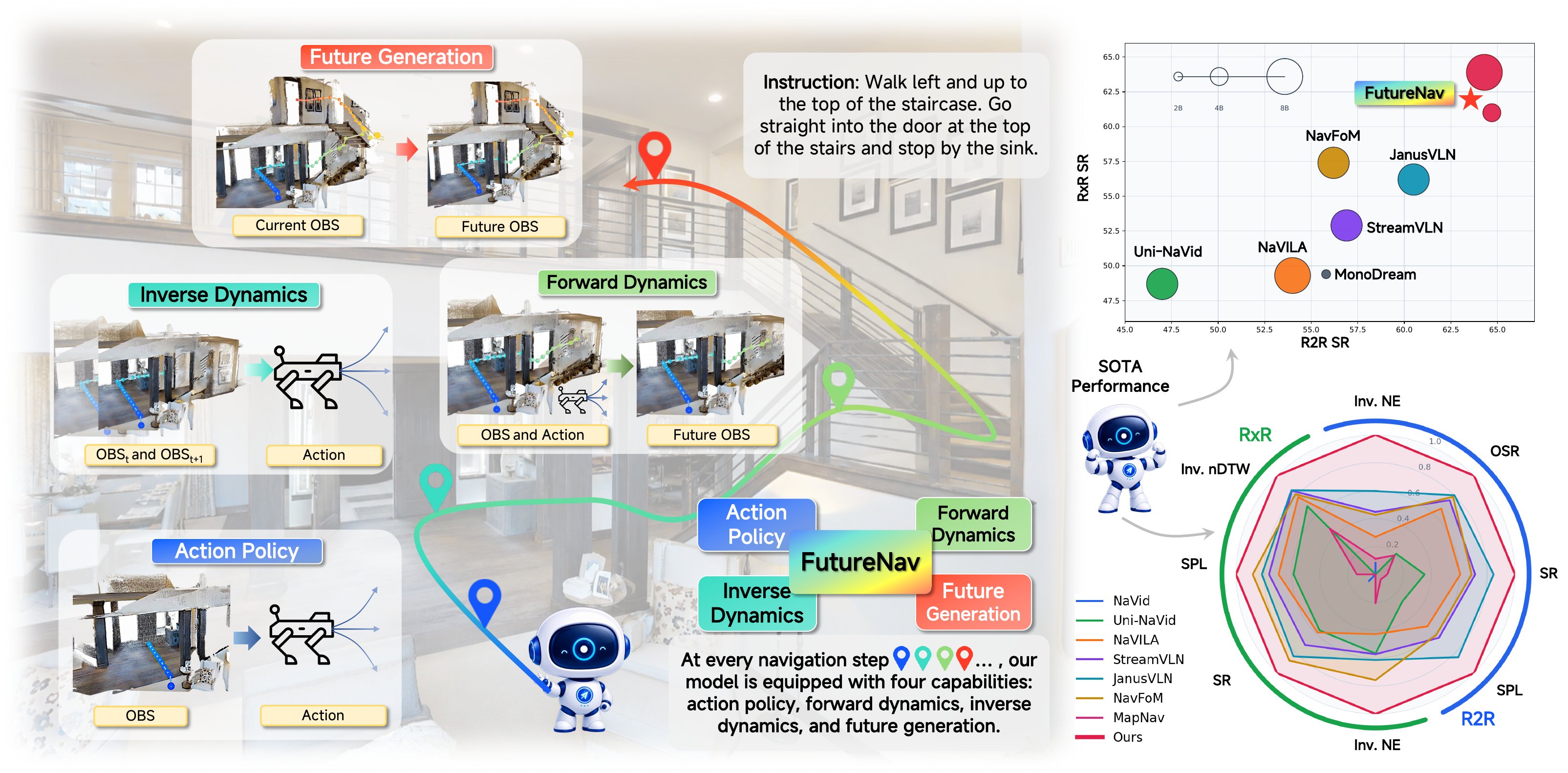}
    \vspace{-2em}
    \caption{Overview of \method{}. Our model unifies four navigation capabilities
    in a single VLM-based world-action framework: action policy learning for direct
    navigation control, inverse dynamics for inferring actions from observation
    changes, forward dynamics for action-conditioned state transition modeling, and
    future generation for predicting upcoming spatial states. This unified design
    improves world understanding and action prediction while achieving
    state-of-the-art performance on VLN-CE benchmarks.}
    \vspace{-1em}
    \label{fig:teaser}
    \end{figure*}

Recent progress has increasingly shifted VLN toward large vision-language models
(VLMs) and vision-language-action foundation models. Methods such as
NaVid~\cite{zhang2024navid}, Uni-NaVid~\cite{zhang2024uni},
NaVILA~\cite{cheng2024navila}, StreamVLN~\cite{wei2025streamvln}, and
Embodied Navigation Foundation Model~\cite{zhang2025embodied} adapt pretrained
multimodal backbones to predict navigation actions from language instructions
and visual histories
\cite{zhang2025nava,cao2025robobrain,hao2025mimoembodied,zhang2026onevla,lu2026xiaomionevl,zhang2025videocot,zhang2025skyready,zhang2025humanoidpano,tang2025roboafford,hao2025roboaffordpp,wu2025evaluatinggpt4o,cheng2025typographic,li2025vquala,hao2026h2rbm,zhang2026whatyoureach,zhang2026meshmimic}.
JanusVLN~\cite{zeng2025janusvln} further separates semantic and spatial
information with dual implicit memories, demonstrating strong performance
without external navigation data. These VLM-based methods provide a scalable
action-decoding interface for VLN, but most of them still formulate navigation
primarily as direct observation-to-action generation. Consequently, they do not
explicitly supervise world-state representation, action-induced state
transition modeling, or future-state prediction before action selection. In
parallel, recent world-model
navigation systems, including Navigation World Models
(NWM)~\cite{bar2025navigation}, WorldVLN~\cite{zhao2026worldvln},
WAM-Nav~\cite{yang2026wamnav}, NavForesee~\cite{liu2025navforesee},
AstraNav-World~\cite{hu2025astranav}, and NavMorph~\cite{yao2025navmorph},
have explored future observation prediction, latent visual foresight, or
world-action modeling for navigation
\cite{liu2026oawam,liu2026thinking,ma2026reasoning,fu2026sefmap,li2026weather}.
However, their world-action modeling remains incomplete: policy learning, action-conditioned dynamics, and future-state generation are often treated as separate components or only partially modeled, rather than being unified within a single navigation policy. As a result, these methods do not fully couple what action to take with how the world state changes and what future state the action may induce. Moreover, their world-modeling modules often remain active during inference, incurring additional computation for future-state imagination, candidate rollouts, or auxiliary planning.

To bridge this gap, we propose \textbf{\emph{FutureNav}}, a VLM-based unified
world-action modeling framework for continuous VLN. As illustrated in
Figure~\ref{fig:teaser},
\textbf{\emph{FutureNav}} augments a VLM backbone with explicit spatial features from a frozen
spatial encoder. Given the instruction and egocentric observation history, the
VLM backbone encodes text and visual tokens, while the spatial encoder extracts
geometry-aware features from the same observations. A lightweight trainable
merger projects the spatial features into the LLM hidden space and fuses them
with visual tokens through residual addition. The LLM therefore receives a
unified multimodal sequence containing text, visual, and spatial features, and
predicts low-level navigation actions autoregressively. This design retains a
single-pass action-decoding inference pathway while giving the policy
explicit access to spatial state information.

\textbf{\emph{FutureNav}} further trains this architecture with four complementary objectives
for simultaneous world and action modeling. First, the action policy follows the
standard VLM decoding interface: after receiving the instruction and
geometry-enhanced visual tokens, the LLM head autoregressively generates the next
navigation action as an assistant response, and is trained with next-token
cross-entropy on action labels such as \texttt{STOP}, \texttt{MOVE\_FORWARD},
\texttt{TURN\_LEFT}, and \texttt{TURN\_RIGHT}. Second, the inverse dynamics head
predicts the intermediate action between two adjacent observations. It takes the
LLM context hidden state together with the mean-pooled visual-token hidden states
of the previous and current observations, as well as their difference, and
classifies the executed action with a cross-entropy loss. Third, the forward
dynamics head models action-conditioned state transition by concatenating the
current-observation hidden state and the ground-truth action-token hidden state
under teacher forcing, and regressing the pooled next spatial feature with an MSE
loss. Finally, the future
generation head predicts the next spatial feature without explicitly using the
action token: it combines the LLM context hidden state with the current visual
hidden state and learns an action-free prior over short-horizon world evolution.
Together, these objectives transform VLM fine-tuning from pure action imitation
into unified world-action learning: the model learns not only which action to
execute, but also the transition and future spatial state implied by that
decision. 
At inference time, our default setting uses only the action policy for direct action decoding, while the dynamics and future-prediction modules can be optionally enabled when explicit world-state estimation or look-ahead reasoning is desired.

Extensive experiments on standard continuous VLN benchmarks demonstrate the
effectiveness and efficiency of \textbf{\emph{FutureNav}}. With only a 4B-scale backbone,
\textbf{\emph{FutureNav}} achieves state-of-the-art performance on multiple VLN benchmarks and
substantially outperforms prior VLN methods, including stronger 7B/8B-scale
navigation models. We further demonstrate strong zero-shot real-world
vision-and-language navigation capability without real-world fine-tuning. Our
contributions are summarized as follows:
\begin{itemize}
    \item We propose \textbf{\emph{FutureNav}}, a VLM-based unified world-action modeling framework for continuous VLN that equips a single LLM action-decoding policy with action prediction, inverse dynamics, forward dynamics, and future spatial-state generation capabilities.

    \item We introduce four complementary world-action training objectives that jointly supervise policy learning, action-induced state transitions, inverse action inference, and future-state prediction, enabling the model to learn both what action to execute and how the spatial state evolves.
    \item We show through extensive experiments that our unified world-action
    modeling significantly improves navigation performance while preserving
    inference efficiency, enabling a 4B model to outperform prior larger VLN
    foundation models.
\end{itemize}

\section{Related Work}
\subsection{Vision-and-Language Navigation}
\begin{figure*}[!ht]
    \centering
    \includegraphics[width=\linewidth]{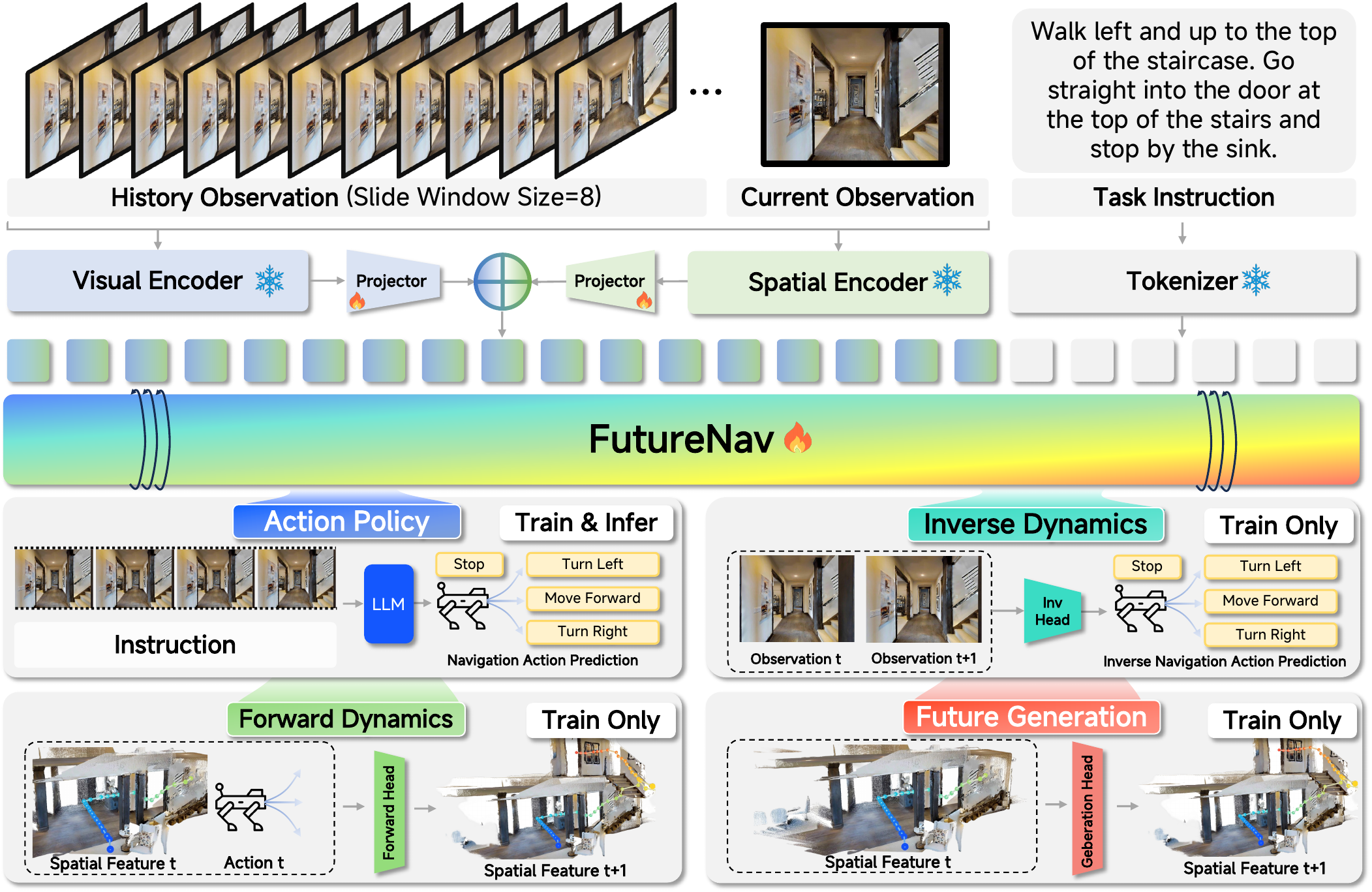}
    \vspace{-1.5em}
    \caption{Architecture of \textbf{\emph{FutureNav}}. A VLM backbone encodes the instruction and egocentric observations, while a frozen spatial encoder provides spatial-aware features for world-state understanding. The model is trained to support four world-action capabilities: action policy learning for navigation control, inverse dynamics for inferring actions from observation changes, forward dynamics for predicting action-conditioned state transitions, and future generation for anticipating upcoming spatial states.}
    \vspace{-1.5em}
    \label{fig:framework}
\end{figure*}

Vision-and-Language Navigation (VLN) requires an embodied agent to follow
language instructions to a goal location. It is commonly evaluated on benchmarks
such as R2R~\cite{anderson2018vision},
RxR~\cite{ku2020room}, and continuous-environment VLN-CE
\cite{krantz2020beyond}. Recent work has shifted toward end-to-end policies
built on VLMs. NaVid~\cite{zhang2024navid} maps a monocular RGB stream and an
instruction to actions without maps, odometry, or depth;
Uni-NaVid~\cite{zhang2024uni} and NaVILA~\cite{cheng2024navila} extend this
paradigm into vision-language-action models for multi-task navigation; and
NavA$^3$~\cite{zhang2025nava} further handles open-ended instructions across diverse
scenes. StreamVLN~\cite{wei2025streamvln}, MapNav~\cite{zhang2025mapnav}, and
JanusVLN~\cite{zeng2025janusvln} target streaming efficiency and long-horizon
memory through slow-fast context modeling, semantic maps, and decoupled implicit
memory, while NavFoM~\cite{zhang2025embodied} scales VLN to a cross-embodiment
Despite this progress, these methods mainly learn a direct mapping from
observations to actions, without explicitly modeling the underlying world state
or how it evolves under the agent's actions. As a result, the policy receives
limited supervision on action-induced state transitions and future spatial
states, which can weaken long-horizon spatial reasoning and decision robustness.
This limitation motivates our unified world-action formulation.

\subsection{World Modeling for Embodied Navigation}

World models predict how an environment evolves in response to actions, enabling
planning through imagined rollouts. In navigation, early world models imagine
future observations to support look-ahead planning:
PathDreamer~\cite{koh2021pathdreamer} synthesizes high-resolution views of unseen
viewpoints, Navigation World Models~\cite{bar2025navigation} learn an
action-conditioned video generator that ranks candidate trajectories, and
VISTAv2~\cite{huang2025vistav2} rolls out egocentric futures into a value map
for planning. Since pixel-level generation is costly and often misaligned with
decision making, recent work has shifted prediction into latent space:
NavMorph~\cite{yao2025navmorph} predicts future visual embeddings to enrich
action prediction, and NavCoT~\cite{lin2025navcot} imagines the next observation
as an intermediate reasoning step. A parallel line further unifies prediction
and action within one model: NavForesee~\cite{liu2025navforesee} couples
language planning with predictive imagination in a single VLM, while
MonoDream~\cite{wang2025monodream} predicts latent panoramic RGB-D features as
an auxiliary objective~\cite{liu2026oawam,liu2026thinking,ma2026reasoning,yuan2026fast,li2026spaact}. Yet these methods either keep the world model separate
from the acting policy, rely on additional depth supervision, or model only a
subset of action-observation relations when unified. In contrast, \textbf{\emph{FutureNav}} jointly learns action policy, forward
dynamics, inverse dynamics, and future generation within a 3D-aware,
instruction-conditioned policy in latent feature space, forming a unified
world-action modeling framework.

\section{Methodology}
\label{sec:method}
\subsection{Overview}

We formulate continuous vision-and-language navigation as conditional action
generation with auxiliary world modeling. At time step $t$, the agent receives a
natural language instruction $x$ and an egocentric observation history
$O_t=\{I_1,\ldots,I_t\}$, where each $I_i$ is an RGB frame. The goal is to predict
the next low-level navigation action
$a_t \in \mathcal{A}$, where
$\mathcal{A}=\{\texttt{STOP},\texttt{MOVE\_FORWARD},\texttt{TURN\_LEFT},
\texttt{TURN\_RIGHT}\}$. We implement the policy as an autoregressive
vision-language model: the instruction, visual placeholders, and previous
context are tokenized into a multimodal sequence, and the model decodes the
action as text.

Figure~\ref{fig:framework} summarizes the overall pipeline. \textbf{\emph{FutureNav}} uses a VLM
backbone as the base vision-language-action model and augments it with a frozen
spatial encoder for spatial state encoding. The VLM visual encoder produces
visual token embeddings from the current observation sequence, while the spatial
encoder processes the same frames into spatial-aware features. A lightweight
trainable merger projects spatial features into the language-model hidden space,
and the projected spatial tokens are fused with VLM visual tokens by residual
addition. The fused tokens, together with text tokens from the instruction and
action prompt, are fed to the VLM language model. On top of the language-model hidden states, we
train four modules: an autoregressive action policy, an action-conditioned
forward dynamics head, an inverse dynamics head, and an action-free future
generation head.

\subsection{Input Encoding}

\textbf{\emph{Text and Visual Tokenization.}}
Each training sample is represented as a chat-style sequence containing the
instruction, one or more \texttt{<image>} placeholders, and the target action
response. The tokenizer converts the instruction and action text into language
tokens, while the image placeholders reserve positions for visual embeddings in
the final input sequence. During supervised training, labels on system and user
tokens are masked with $\texttt{-100}$, so the language-model loss is applied
only to assistant-side action tokens.

\textbf{\emph{Visual Encoder.}}
For every RGB observation frame, the VLM image processor produces
pixel-level inputs and the corresponding spatial grid metadata. These inputs are
fed into the VLM vision encoder, yielding visual embeddings
$E^{q}_t$ and deep visual features used by the language model. When an
observation history contains multiple images, all frames are processed and their
visual tokens are placed at the corresponding \texttt{<image>} positions in the
multimodal sequence.

\textbf{\emph{Spatial Encoder.}}
In parallel, the same observation frames are loaded with the spatial-encoder
preprocessing pipeline and passed through a frozen spatial encoder, implemented
with VGGT \cite{wang2025vggt}. The spatial encoder is never updated during
training. For each frame, the encoder consumes the current image along with its
cached history and outputs spatial tokens from an intermediate transformer layer.
We use these tokens as spatial-aware features
$G_t \in \mathbb{R}^{N_g \times 2048}$, where $N_g$ is the number of spatial
patches. The implementation groups neighboring patches according to the VLM
spatial merge size, applies an RMS normalization, and maps the grouped
spatial features to the VLM hidden size with a two-layer MLP:
\begin{equation}
    E^{g}_t = \mathrm{Merger}_{\theta}(G_t).
\end{equation}
If the number of projected spatial tokens differs from the VLM visual token
count, we linearly interpolate $E^{g}_t$ along the token dimension for alignment.
The final visual embedding inserted into the language-model input is
\begin{equation}
    E^{v}_t = E^{q}_t + \alpha E^{g}_t,
\end{equation}
where $\alpha$ is a scalar residual weight. This fusion keeps the pretrained
VLM visual representation intact while injecting explicit spatial-aware
features into the same token stream consumed by the language model.

\begin{figure*}[!ht]
    \centering
    \includegraphics[width=\linewidth]{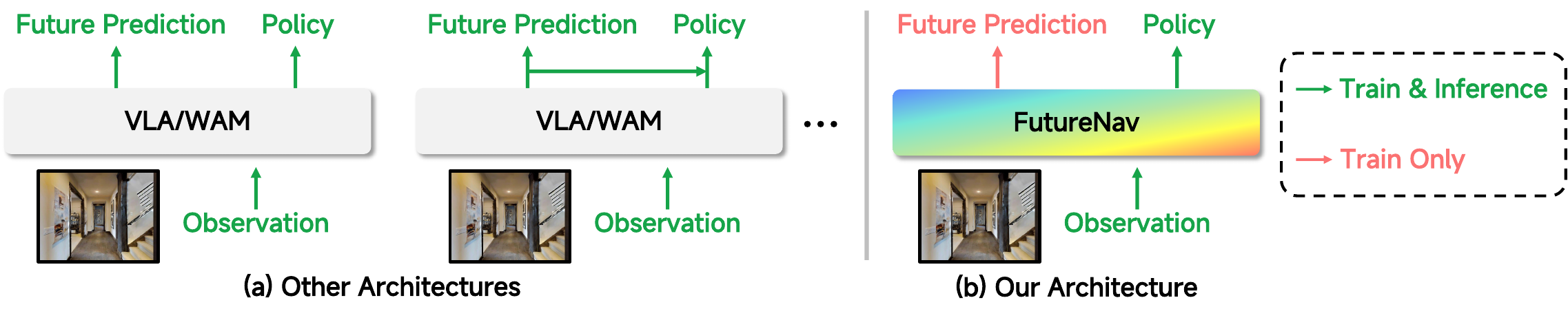}
    \vspace{-2em}
    \caption{Inference comparison between \textbf{\emph{FutureNav}} and existing
world-model-based navigation methods. Prior methods typically predict future
states together with the policy or use future prediction to guide action
selection at inference time, introducing additional computation. In contrast,
\textbf{\emph{FutureNav}} separates policy decoding from world-state prediction:
the model learns policy, inverse dynamics, forward dynamics, and future
generation during training, while inference can use only the policy branch for
fast action prediction.}
\vspace{-1em}
    \label{fig:training-objective}
\end{figure*}

\subsection{World-Action Modeling}

\textbf{\emph{Action Policy.}}
The main navigation policy is the VLM language model conditioned on the
spatial-enhanced multimodal sequence. Let $H_t=\{h_1,\ldots,h_L\}$ denote the
hidden states output by the language model. A tied language-model head projects
each hidden state to vocabulary logits, and the action is decoded
autoregressively as an assistant response. In our implementation, each low-level
navigation action is represented as text, including \texttt{STOP},
\texttt{MOVE\_FORWARD}, \texttt{TURN\_LEFT}, and \texttt{TURN\_RIGHT}. The
action loss is the standard next-token cross entropy over the unmasked assistant
action tokens:
\begin{equation}
    \mathcal{L}_{\mathrm{policy}}
    = - \sum_{\ell \in \mathcal{Y}} \log p_{\theta}(y_{\ell}\mid y_{<\ell}, x, O_t),
\end{equation}
where $\mathcal{Y}$ denotes unmasked assistant action tokens. This objective
teaches the model to map instructions and observations to executable navigation
actions.

\textbf{\emph{Forward Dynamics.}}
The forward dynamics head models action-conditioned state transition. During
training, the action tokens are available under teacher forcing. We take the
hidden state at the last image-token position as the current observation state
$h^{\mathrm{obs}}_t$ and the hidden state at the last unmasked ground-truth
action-token position as the action state $h^{\mathrm{act}}_t$. The two states
are concatenated and passed through a lightweight MLP:
\begin{equation}
    \hat{z}_{t+1}
    =
    f_{\theta}\left([h^{\mathrm{obs}}_t;h^{\mathrm{act}}_t]\right),
    \quad
    \hat{z}_{t+1} \in \mathbb{R}^{2048}.
\end{equation}
The target $z_{t+1}$ is the pooled spatial-encoder feature of  next
observation, represented as a latent vector. The forward
dynamics loss is
\begin{equation}
    \mathcal{L}_{\mathrm{forward}}
    = \left\|\hat{z}_{t+1}-z_{t+1}\right\|_2^2.
\end{equation}
This module explicitly asks the model to predict what it will observe after
executing the selected action.

\textbf{\emph{Inverse Dynamics.}}
The inverse dynamics head encourages visual representations to encode
action-relevant changes between adjacent observations. We first split the
language-model hidden states at image-token positions into per-frame groups using
the image grid metadata. The previous and current observation states are obtained
by mean-pooling the last two image-token groups:
\begin{equation}
    \bar{h}^{v}_{t-1}=\mathrm{MeanPool}(H^{v}_{t-1}), \quad
    \bar{h}^{v}_{t}=\mathrm{MeanPool}(H^{v}_{t}).
\end{equation}
We also use a context hidden state $h^{\mathrm{ctx}}_t$, taken from the token
immediately before assistant action response, and an explicit visual-change
feature $\Delta h^{v}_t=\bar{h}^{v}_{t}-\bar{h}^{v}_{t-1}$. The inverse head
classifies the action that transforms the previous observation into current
one:
\begin{equation}
    \hat{a}_t =
    g_{\theta}\left([h^{\mathrm{ctx}}_t;\bar{h}^{v}_{t-1};\bar{h}^{v}_{t};\Delta h^{v}_t]\right),
    \quad
    \hat{a}_t \in \mathbb{R}^{4}.
\end{equation}
The classifier predicts one of the four discrete actions in $\mathcal{A}$ and is
supervised by the previous-action label:
\begin{equation}
    \mathcal{L}_{\mathrm{inverse}}
    =
    \mathrm{CE}(\hat{a}_t, a_t).
\end{equation}
Unlike forward head, which predicts future spatial state from an action, 
inverse head recovers action from observed state change. The two objectives therefore
provide complementary constraints on action-state consistency.

\textbf{\emph{Future Generation.}}
The generation head predicts future spatial state without using the action
token. Unlike the forward head, it does not condition on the teacher-forced
action representation. Instead, it combines the context hidden state
$h^{\mathrm{ctx}}_t$ with the mean-pooled current image-token hidden state
$\bar{h}^{v}_{t}$ and predicts the next spatial feature:
\begin{equation}
    \hat{z}^{\mathrm{gen}}_{t+1}
    =
    r_{\theta}\left([h^{\mathrm{ctx}}_t;\bar{h}^{v}_{t}]\right),
    \quad
    \hat{z}^{\mathrm{gen}}_{t+1} \in \mathbb{R}^{2048}.
\end{equation}
It uses the same pooled next-observation spatial feature $z_{t+1}$ as the
prediction target and is trained with an MSE loss:
\begin{equation}
    \mathcal{L}_{\mathrm{gen}}
    =
    \left\|\hat{z}^{\mathrm{gen}}_{t+1}-z_{t+1}\right\|_2^2.
\end{equation}
Because this head does not receive the action token, it complements the forward
dynamics head by learning an action-agnostic prior over short-horizon visual
evolution and scene continuity.

\subsection{Training Objective}

The final training objective combines action generation with the three auxiliary
world-modeling losses:
\begin{equation}
    \mathcal{L}
    =
    \mathcal{L}_{\mathrm{policy}}
    + \lambda_f \mathcal{L}_{\mathrm{forward}}
    + \lambda_i \mathcal{L}_{\mathrm{inverse}}
    + \lambda_g \mathcal{L}_{\mathrm{gen}}.
\end{equation}
Only VLM language components, the multimodal merger, and the world-action modeling heads are
optimized; the visual and spatial encoder remains frozen. As illustrated in
Figure~\ref{fig:training-objective}, the auxiliary heads shape the shared
representation during training but are not required for evaluation. In our main
setting, inference uses only the action policy branch for direct action decoding,
so world modeling introduces no additional inference cost.

\section{Experiments}
\subsection{Experimental Setup}

\begin{table*}[!ht]
    \centering
    \definecolor{futurenavrow}{HTML}{F2E7F5}
    \definecolor{futurenavdelta}{HTML}{3B0047}
    \fontsize{16}{24}\selectfont
    \setlength{\tabcolsep}{2pt}
    \renewcommand{\arraystretch}{1.02}
    \caption{Comparison with prior VLN-CE methods on the R2R-CE and RxR-CE val-unseen splits. \textbf{\emph{FutureNav}} uses only a single RGB sensor. StreamVLN* uses EnvDrop as external data. NaVILA* excludes human-following data. $^\dag$ denotes models trained only with VLN-CE expert trajectories and without any additional data. $\Delta$(\%) rows report relative gains over JanusVLN; NE is computed in the reverse direction since lower is better.}
    \label{tab:r2r_rxr_sota}
    \resizebox{\textwidth}{!}{
    \begin{tabular}{lcccccccccccccc}
    \toprule
    \multirow{2}{*}{Method} & \multirow{2}{*}{\shortstack{VLM\\Params}} & \multicolumn{4}{c}{Observation} & \multicolumn{4}{c}{R2R Val-Unseen} & \multicolumn{4}{c}{RxR Val-Unseen} & \multirow{2}{*}{\shortstack{External\\Training\\Data}} \\
    \cmidrule(lr){3-6}\cmidrule(lr){7-10}\cmidrule(lr){11-14}
    & & Pano. & Odo. & Depth & S.RGB & NE$\downarrow$ & OS$\uparrow$ & SR$\uparrow$ & SPL$\uparrow$ & NE$\downarrow$ & SR$\uparrow$ & SPL$\uparrow$ & nDTW$\uparrow$ & \\
    \midrule

    HPN+DN$_{\text{[ICCV'21]}}$~\cite{krantz2021waypoint}          & -  & $\checkmark$ & $\checkmark$ & $\checkmark$ &              & 6.31 & 40.0 & 36.0 & 34.0 & -    & -    & -    & - & - \\
    CMA$_{\text{[CVPR'22]}}$~\cite{hong2022bridging}             & -  & $\checkmark$ & $\checkmark$ & $\checkmark$ &              & 6.20 & 52.0 & 41.0 & 36.0 & 8.76 & 26.5 & 22.1 & 47.0 & - \\
    Sim2Sim$_{\text{[ECCV'22]}}$~\cite{krantz2022sim}         & -  & $\checkmark$ & $\checkmark$ & $\checkmark$ &              & 6.07 & 52.0 & 43.0 & 36.0 & -    & -    & -    & - & - \\
    VLN$\circlearrowright$BERT$_{\text{[CVPR'22]}}$~\cite{hong2022bridging}        & -  & $\checkmark$ & $\checkmark$ & $\checkmark$ &              & 5.74 & 53.0 & 44.0 & 39.0 & 8.98 & 27.0 & 22.6 & 46.7 & - \\
    Ego$^2$-Map$_{\text{[ICCV'23]}}$~\cite{hong2023learning}     & -  & $\checkmark$ & $\checkmark$ & $\checkmark$ &              & 5.54 & 56.0 & 47.0 & 41.0 & -    & -    & -    & - & - \\
    DreamWalker$_{\text{[ICCV'23]}}$~\cite{wang2023dreamwalker}     & -  & $\checkmark$ & $\checkmark$ & $\checkmark$ &              & 5.53 & 59.0 & 49.0 & 44.0 & -    & -    & -    & - & - \\
    GridMM$_{\text{[ICCV'23]}}$~\cite{wang2023gridmm}          & -  & $\checkmark$ & $\checkmark$ & $\checkmark$ &              & 5.11 & 61.0 & 49.0 & 41.0 & -    & -    & -    & - & - \\
    Reborn$_{\text{[ICCV'23]}}$~\cite{an20221st}          & -  & $\checkmark$ & $\checkmark$ & $\checkmark$ &              & 5.40 & 57.0 & 50.0 & 46.0 & 5.98 & 48.6 & 42.0 & 63.3 & - \\
    InstructNav$_{\text{[CoRL'24]}}$~\cite{long2024instructnav}     & -  & $\checkmark$ & $\checkmark$ & $\checkmark$ &              & 6.89 & -    & 31.0 & 24.0 & -    & -    & -    & - & - \\
    \midrule
    COSMO$_{\text{[ICCV'25]}}$~\cite{zhang2025cosmo}           & -  & $\checkmark$ &              &              &              & -    & 56.0 & 47.0 & 40.0 & -    & -    & -    & - & - \\
    AO-Planner$_{\text{[AAAI'25]}}$~\cite{chen2025affordances}      & -  & $\checkmark$ &              & $\checkmark$ &              & 5.55 & 59.0 & 47.0 & 33.0 & 7.06 & 43.3 & 30.5 & 50.1 & - \\
    LAW$_{\text{[EMNLP'21]}}$~\cite{raychaudhuri2021language}            & -  &              & $\checkmark$ & $\checkmark$ & $\checkmark$ & 6.83 & 44.0 & 35.0 & 31.0 & 10.90 & 8.0 & 8.0 & 38.0 & - \\
    MapNav$_{\text{[ACL'25]}}$~\cite{zhang2025mapnav}          & -  &              & $\checkmark$ & $\checkmark$ & $\checkmark$ & 4.93 & 53.0 & 39.7 & 37.2 & -    & -    & -    & - & - \\
    g3D-LF$_{\text{[CVPR'25]}}$~\cite{wang2025g3d}          & -  &              & $\checkmark$ & $\checkmark$ & $\checkmark$ & 5.70 & 59.5 & 47.2 & 34.6 & -    & -    & -    & - & - \\
    Seq2Seq$_{\text{[ECCV'20]}}$~\cite{krantz2020beyond}         & -  &              &              & $\checkmark$ & $\checkmark$ & 7.77 & 37.0 & 25.0 & 22.0 & 12.10 & 13.9 & 11.9 & 30.8 & - \\
    NaVid-4D$_{\text{[ICRA'25]}}$~\cite{liu2025vid}        & -  &              &              & $\checkmark$ & $\checkmark$ & 5.99 & 55.7 & 43.8 & 37.1 & -    & -    & -    & - & - \\
    NavMorph$_{\text{[ICCV'25]}}$~\cite{yao2025navmorph}        & -  &              &              & $\checkmark$ & $\checkmark$ & 5.75 & 56.9 & 47.9 & 33.2 & 8.85 & 30.8 & 22.8 & 44.2 & - \\
    \midrule
    NaVid$_{\text{[RSS'24]}}$~\cite{zhang2024navid}            & 7B &              &              &              & $\checkmark$ & 5.47 & 49.1 & 37.4 & 35.9 & -    & -    & -    & - & 953K \\
    Sim2Real$_{\text{[CoRL'24]}}$~\cite{wang2024sim}        & -  &              &              &              & $\checkmark$ & 5.95 & 55.8 & 44.9 & 30.4 & 8.79 & 36.7 & 25.5 & 18.1 & 0K \\
    Uni-NaVid$_{\text{[RSS'25]}}$~\cite{zhang2024uni}        & 7B &              &              &              & $\checkmark$ & 5.58 & 53.3 & 47.0 & 42.7 & 6.24 & 48.7 & 40.9 & - & 3577K \\
    InternVLA-N1$_{\text{[arXiv'25]}}$~\cite{wei2025ground} & 8B &              &              &              & $\checkmark$ & 4.89 & 60.6 & 55.4 & 52.1 & 6.41 & 49.5 & 41.8 & 62.6 & 51200K \\
    StreamVLN*$_{\text{[ICRA'26]}}$~\cite{wei2025streamvln}     & 7B &              &              &              & $\checkmark$ & 6.05 & 53.8 & 45.5 & 41.6 & -    & -    & -    & - & 10033K \\
    NaVILA*$_{\text{[RSS'25]}}$~\cite{cheng2024navila}          & 7B &              &              &              & $\checkmark$ & 5.37 & 57.6 & 49.7 & 45.5 & -    & -    & -    & - & 12574K \\
    NaVILA$_{\text{[RSS'25]}}$~\cite{cheng2024navila}           & 7B &              &              &              & $\checkmark$ & 5.22 & 62.5 & 54.0 & 49.0 & 6.77 & 49.3 & 44.0 & 58.8 & 13132K \\
    JanusVLN$^\dag$$_{\text{[ICLR'26]}}$~\cite{zeng2025janusvln}        & 7B &              &              &              & $\checkmark$ & 5.17 & 58.0 & 52.8 & 49.2 & 6.46 & 51.4 & 44.3 & 59.1 & 0K \\
    \rowcolor{futurenavrow}
    \textbf{\emph{FutureNav}-4B$^\dag$ (Ours)}    & \textbf{4B} &              &              &              & $\checkmark$ & \textbf{5.13} & \textbf{61.8} & \textbf{55.1} & \textbf{50.1} & \textbf{5.62} & \textbf{54.5} & \textbf{46.0} & \textbf{61.8} & \textbf{0K} \\

    \rowcolor{futurenavrow}
    \textbf{\emph{FutureNav}-8B$^\dag$ (Ours)}    & \textbf{8B} &              &              &              & $\checkmark$ & \textbf{5.15} & \textbf{61.6} & \textbf{55.5} & \textbf{51.4} & \textbf{5.93} & \textbf{53.3} & \textbf{45.4} & \textbf{59.8} & \textbf{0K} \\

    StreamVLN$_{\text{[ICRA'26]}}$~\cite{wei2025streamvln}      & 7B &              &              &              & $\checkmark$ & 4.98 & 64.2 & 56.9 & 51.9 & 6.22 & 52.9 & 46.0 & 61.9 & $\sim$26330K \\
    NavFoM$_{\text{[ICLR'26]}}$~\cite{zhang2025embodied}        & 7B &              &              &              & $\checkmark$ & 5.01 & 64.9 & 56.2 & 51.2 & 5.51 & 57.4 & 49.4 & 60.2 & $\sim$10480K \\
    JanusVLN$_{\text{[ICLR'26]}}$~\cite{zeng2025janusvln}         & 7B &              &              &              & $\checkmark$ & 4.78 & 65.2 & 60.5 & 56.8 & 6.06 & 56.2 & 47.5 & 62.1 & 10692K \\
    \rowcolor{futurenavrow}
    \textbf{\emph{FutureNav}-4B (Ours)}    & \textbf{4B} &              &              &              & $\checkmark$ & \textbf{4.24} & \textbf{70.2} & \textbf{65.4} & \textbf{61.3} & \textbf{4.60} & \textbf{60.9} & \textbf{52.8} & \textbf{68.0} & \textbf{10457K} \\
    \textcolor{futurenavdelta}{\itshape $\Delta$(\%) vs. JanusVLN} & & & & & & \textcolor{futurenavdelta}{\itshape +11.3} & \textcolor{futurenavdelta}{\itshape +7.7} & \textcolor{futurenavdelta}{\itshape +8.1} & \textcolor{futurenavdelta}{\itshape +7.9} & \textcolor{futurenavdelta}{\itshape +24.1} & \textcolor{futurenavdelta}{\itshape +8.4} & \textcolor{futurenavdelta}{\itshape +11.2} & \textcolor{futurenavdelta}{\itshape +9.5} & \\
    \rowcolor{futurenavrow}
    \textbf{\emph{FutureNav}-8B (Ours)}    & \textbf{8B} &              &              &              & $\checkmark$ & \textbf{4.29} & \textbf{69.2} & \textbf{64.3} & \textbf{60.3} & \textbf{4.26} & \textbf{63.9} & \textbf{54.8} & \textbf{69.1} & \textbf{10435K} \\
    \textcolor{futurenavdelta}{\itshape $\Delta$(\%) vs. JanusVLN} & & & & & & \textcolor{futurenavdelta}{\itshape +10.3} & \textcolor{futurenavdelta}{\itshape +6.1} & \textcolor{futurenavdelta}{\itshape +6.3} & \textcolor{futurenavdelta}{\itshape +6.2} & \textcolor{futurenavdelta}{\itshape +29.7} & \textcolor{futurenavdelta}{\itshape +13.7} & \textcolor{futurenavdelta}{\itshape +15.4} & \textcolor{futurenavdelta}{\itshape +11.3} & \\
    \bottomrule
    \end{tabular}
    }
    \vspace{-1em}
    \end{table*}

\textbf{Dataset and Evaluation Metrics.}
We evaluate \textbf{\emph{FutureNav}} on two standard continuous VLN
benchmarks, R2R-CE~\cite{krantz2020beyond} and RxR-CE~\cite{ku2020room}. Both datasets instantiate instruction-following
navigation in the Habitat simulator \cite{savva2019habitat} with high-fidelity
Matterport3D indoor scans \cite{chang2017matterport3d}. 
Following prior continuous VLN work
\cite{cheng2024navila,wei2025streamvln,zeng2025janusvln}, we report results on
the validation-unseen splits, which contain 1,839 trajectories for R2R-CE and
3,669 trajectories for RxR-CE. We report the standard VLN-CE metrics, including
Navigation Error (NE), Oracle Success (OS), Success Rate (SR), and Success
weighted by Path Length (SPL), and additionally report normalized Dynamic Time
Warping (nDTW) on RxR-CE. We treat SR and SPL as the primary metrics.

\textbf{Implementation Details.}
\textbf{\emph{FutureNav}} instantiates the VLM backbone with Qwen3-VL-4B and
Qwen3-VL-8B~\cite{bai2025qwen3vltechnicalreport}, together with a frozen spatial
encoder for spatial-aware features and future-state targets. During navigation
fine-tuning, we optimize the language model, multimodal merger, and four
world-action heads, while keeping the VLM vision tower and spatial encoder
frozen. The residual spatial fusion weight is set to $\alpha=0.2$. In the 0K
setting, models are trained only with expert trajectories from R2R-CE and
RxR-CE, which provide approximately 2.43M training samples, including 631K from
R2R-CE and 1.80M from RxR-CE. In the full-data setting, we further add external
navigation data from ScaleVLN~\cite{wang2023scaling} and DAgger
rollouts~\cite{ross2011reduction}.Unless otherwise specified, the policy loss weight is set to $1.0$, and the auxiliary world-action loss weights are set to $\lambda_f=\lambda_i=\lambda_g=0.1$.

\textbf{Real-world Deployment.}
For real-world experiments, we deploy \textbf{\emph{FutureNav}} on an H20 GPU
server for inference and use a remotely operated Go2 robot as the physical
navigation platform. The robot is equipped with a D435i camera, which captures
egocentric RGB observations and streams them to the inference server. At each
navigation step, the server predicts a low-level action from the language
instruction and the received RGB history, and the Go2 robot executes the
returned action in the physical environment. We evaluate real-world navigation
in a zero-shot setting without additional real-world fine-tuning.

\begin{figure*}[!ht]
    \centering
    \includegraphics[width=\linewidth]{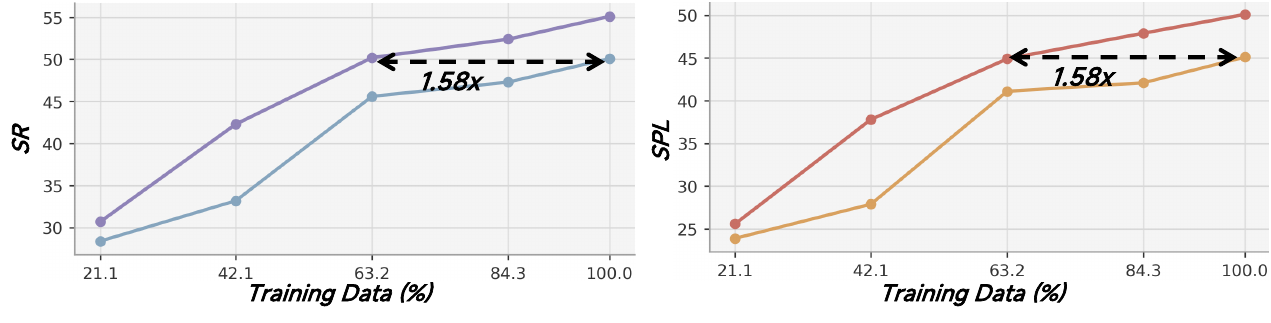}
    \vspace{-2.5em}
    \caption{Training efficiency ablation on the R2R-CE val-unseen split. We compare \textbf{\emph{FutureNav}} with a policy-only baseline under matched training progress and report SR and SPL as functions of the completed training steps. \textbf{\emph{FutureNav}} improves faster throughout training and maintains higher final navigation accuracy.}
    \label{fig:training_efficiency}
\end{figure*}

\begin{table*}[t]
\centering
\scriptsize
\definecolor{futurenavrow}{HTML}{F2E7F5}
\begin{minipage}[t]{0.46\textwidth}
\centering
\setlength{\tabcolsep}{3.5pt}
\renewcommand{\arraystretch}{1.05}
\caption{Component ablation on the R2R-CE val-unseen split under the 0K setting.}
\vspace{-1em}
\label{tab:component_ablation_r2r}
\resizebox{\linewidth}{!}{
\begin{tabular}{cccccccc}
\toprule
Policy & Inv. & Fwd. & Gen. & NE$\downarrow$ & OS$\uparrow$ & SR$\uparrow$ & SPL$\uparrow$ \\
\midrule
$\checkmark$ &              &              &              & 6.00 & 56.5 & 50.1 & 45.1 \\
$\checkmark$ & $\checkmark$ &              &              & 5.82 & 57.4 & 50.8 & 45.8 \\
$\checkmark$ &              & $\checkmark$ &              & 5.31 & 59.8 & 53.0 & 48.3 \\
$\checkmark$ &              &              & $\checkmark$ & 5.35 & 58.3 & 51.6 & 46.9 \\
\rowcolor{futurenavrow}
$\checkmark$ & $\checkmark$ & $\checkmark$ & $\checkmark$ & \textbf{5.13} & \textbf{61.8} & \textbf{55.1} & \textbf{50.1} \\
\bottomrule
\end{tabular}
}
\end{minipage}
\hfill
\begin{minipage}[t]{0.46\textwidth}
\centering
\setlength{\tabcolsep}{3.5pt}
\renewcommand{\arraystretch}{1.05}
\caption{Ablation of latent representation on the R2R-CE val-unseen split without external data.}
\vspace{-1em}
\label{tab:latent_prediction_ablation_r2r}
\resizebox{\linewidth}{!}{
\begin{tabular}{lcccc}
\toprule
Latent Representation & NE$\downarrow$ & OS$\uparrow$ & SR$\uparrow$ & SPL$\uparrow$ \\
\midrule
No latent modeling & 6.00 & 56.5 & 50.1 & 45.1 \\
VAE latent & 5.43 & 57.8 & 51.0 & 46.4 \\
VLM visual latent & 5.39 & 58.1 & 51.4 & 46.8 \\
DINO semantic latent & 5.34 & 58.6 & 51.8 & 47.1 \\
\rowcolor{futurenavrow}
\textbf{\emph{Spatial latent (Ours)}} & \textbf{5.13} & \textbf{61.8} & \textbf{55.1} & \textbf{50.1} \\
\bottomrule
\end{tabular}
}
\end{minipage}
\vspace{-1em}
\end{table*}

\subsection{Main Results}

Table~\ref{tab:r2r_rxr_sota} compares \textbf{\emph{FutureNav}} with prior
VLN-CE methods and recent single-RGB VLM navigation models. Under the $^\dag$
setting, where models are trained only with VLN-CE expert trajectories and no
external data, \textbf{\emph{FutureNav}} already achieves strong performance.
\textbf{\emph{FutureNav}}-4B$^\dag$ improves over JanusVLN$^\dag$ by
\emph{4.4\%} in R2R SR and \emph{6.0\%} in RxR SR, while
\textbf{\emph{FutureNav}}-8B$^\dag$ further
reaches 55.5 SR and 51.4 SPL on R2R-CE. Notably, these 0K models already
outperform many recent SOTA methods, including Uni-NaVid, StreamVLN*, NaVILA*,
and NaVILA on R2R SR/SPL. On RxR-CE, the $^\dag$ variants also remain
competitive with previous single-RGB navigation models despite using no external
training data. This shows that the improvement
does not simply come from additional training data.

With the full training recipe, \textbf{\emph{FutureNav}} uses an external-data
budget comparable to JanusVLN and NavFoM, and smaller than StreamVLN and
InternVLA-N1. Under this setting, both 4B and 8B checkpoints surpass all prior
single-RGB SOTA methods. On R2R-CE, \textbf{\emph{FutureNav}}-4B achieves the
best overall results, improving over JanusVLN by \emph{11.3\%} in NE,
\emph{8.1\%} in SR, and \emph{7.9\%} in SPL, despite using slightly less
external data (10.457M vs. 10.692M).
On RxR-CE, \textbf{\emph{FutureNav}}-8B gives the strongest performance,
reducing NE by \emph{29.7\%} over JanusVLN and improving SR/SPL/nDTW by
\emph{13.7\%}/\emph{15.4\%}/\emph{11.3\%}. Compared with the strongest
previous RxR single-RGB results from NavFoM and InternVLA-N1, it also reduces NE
by \emph{22.7\%} and improves SR, SPL, and nDTW by \emph{11.3\%},
\emph{10.9\%}, and \emph{10.4\%}, respectively. These results indicate that
the unified world-action modeling design improves both data efficiency and final
navigation accuracy.

\subsection{Ablation Studies}

\textbf{Effect of World-Action Objectives.}
Table~\ref{tab:component_ablation_r2r} studies the contribution of each
world-action objective under the 0K setting, where all variants are trained only
with VLN-CE expert trajectories. The policy-only baseline reaches 50.1 SR and
45.1 SPL, showing that direct action supervision alone already provides a
competitive navigation policy. Adding a single world-action objective
consistently improves performance. Forward dynamics gives the largest
individual gain, increasing SR/SPL to 53.0/48.3, while future generation further
improves the baseline to 51.6/46.9. Inverse dynamics brings a smaller but still
positive improvement, raising SR/SPL to 50.8/45.8. Combining all three
objectives gives the best result, improving over the policy-only baseline by
5.0 SR and 5.0 SPL and reducing NE from 6.00 to 5.13. These results indicate
that the world-action objectives provide complementary supervision beyond action
imitation.

\textbf{Effect of Spatial Latent Representation.}
Table~\ref{tab:latent_prediction_ablation_r2r} compares different latent
representations for world modeling under the 0K setting. Using VAE
latents~\cite{kingma2014autoencoding} gives 51.0 SR and 46.4 SPL, while VLM
visual latents slightly improve the result to 51.4 SR and 46.8 SPL. DINO
semantic latents~\cite{caron2021emerging} perform better, reaching 51.8 SR and
47.1 SPL, suggesting that semantic visual representations provide more useful
supervision than generic reconstruction latents or backbone visual features in
this setting. Our proposed spatial latent representation performs best,
reaching 55.1 SR and 50.1 SPL with the lowest NE of 5.13, demonstrating the
benefit of explicitly modeling spatial-aware latent states for future world
modeling.

\begin{figure*}[!ht]
    \centering
    \includegraphics[width=\linewidth]{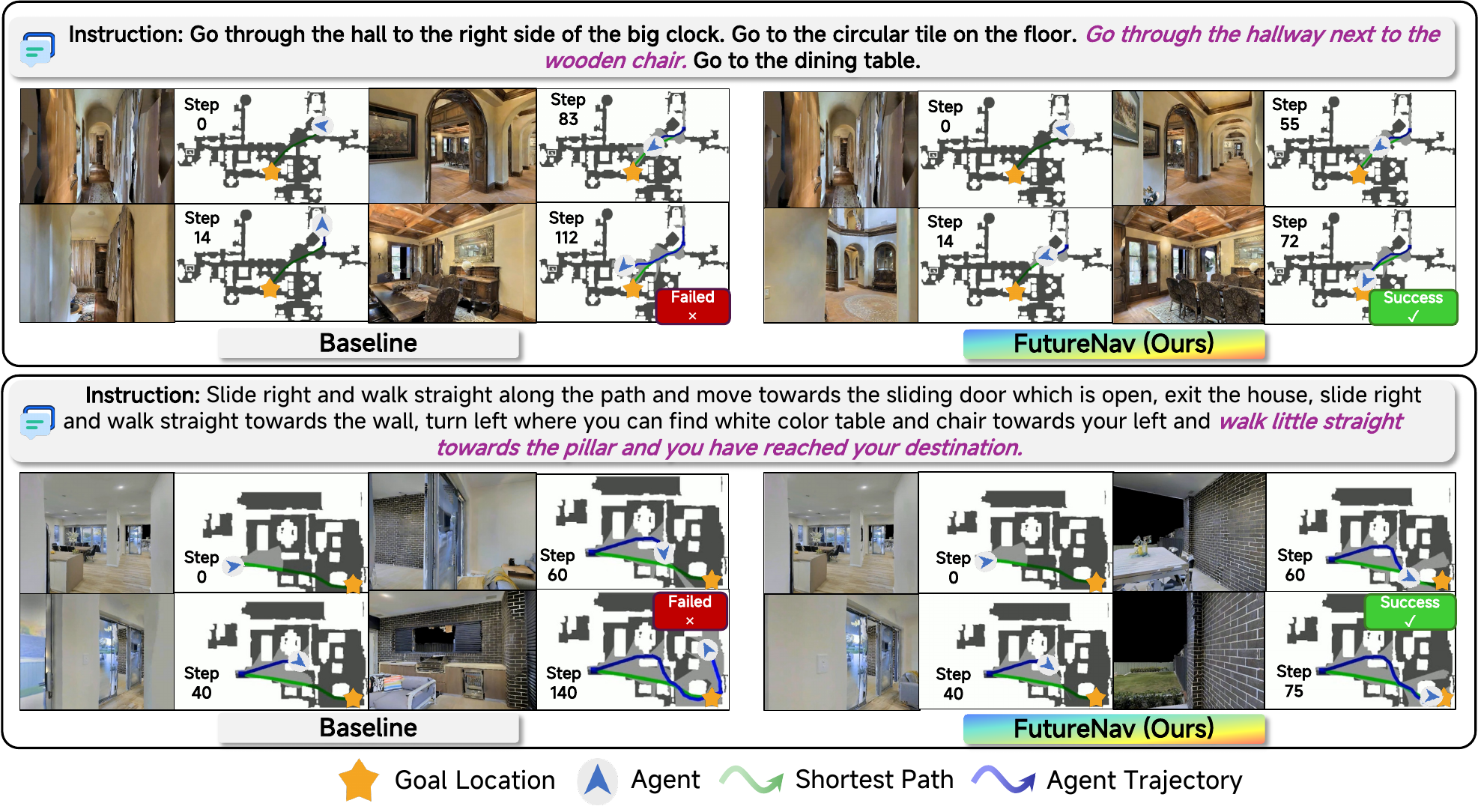}
    \vspace{-2em}
    \caption{Simulated qualitative comparison on VLN-CE. We compare \textbf{\emph{FutureNav}} with JanusVLN~\cite{zeng2025janusvln} under matched instructions and observations. JanusVLN tends to make locally plausible but less stable decisions, while \textbf{\emph{FutureNav}} better preserves spatial progress, follows landmark-conditioned turns, and stops near the intended target.}
    \vspace{-1em}
    \label{fig:qualitative_sim}
\end{figure*}

\begin{figure*}[!ht]
    \centering
    \includegraphics[width=\linewidth]{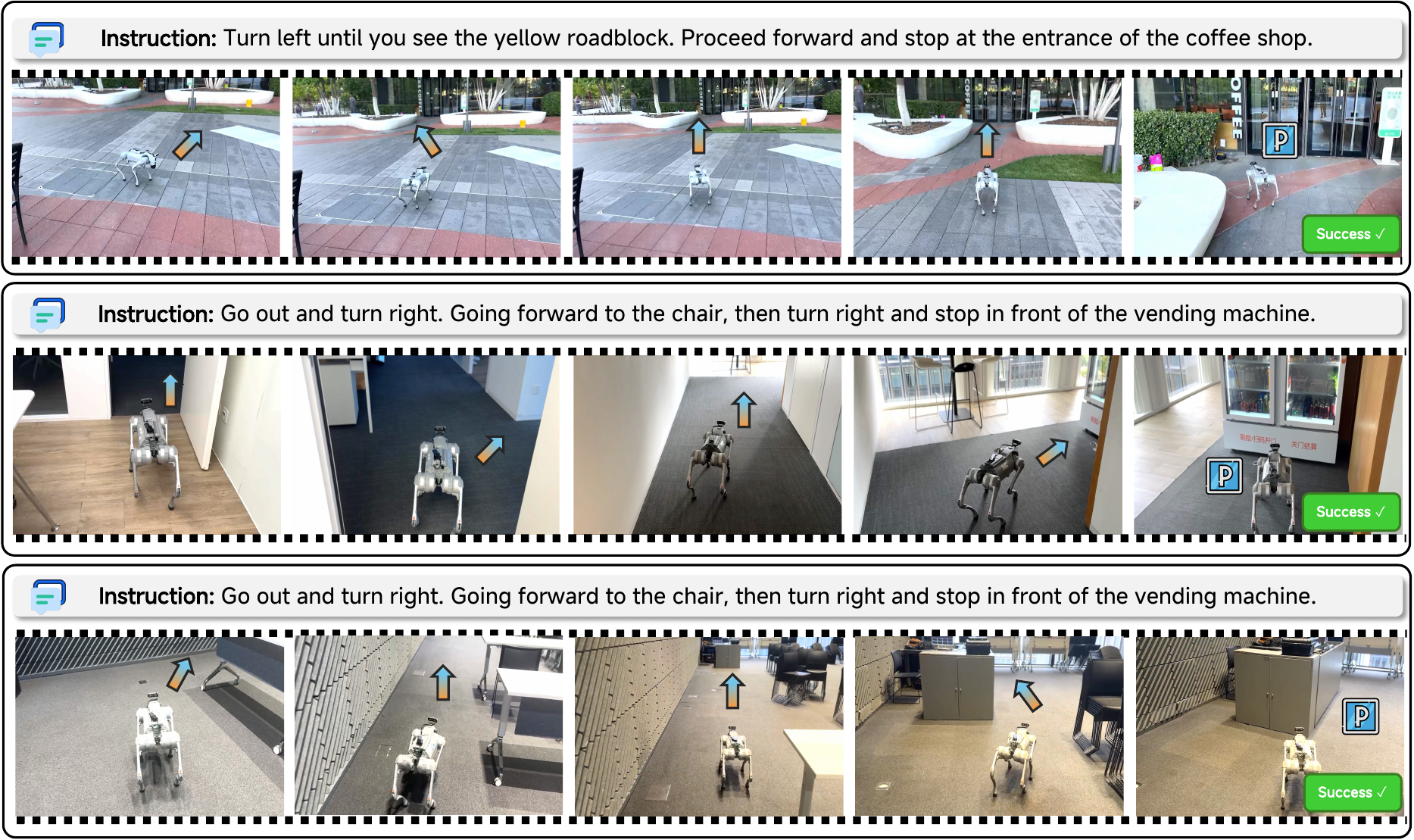}
    \vspace{-1.5em}
    \caption{Zero-shot real-world qualitative results of \textbf{\emph{FutureNav}}. Without real-world navigation fine-tuning, the model follows natural language instructions in real indoor and outdoor scenes, grounds navigation decisions to visible landmarks, and stops at the specified destination.}
    \vspace{-1em}
    \label{fig:qualitative}
\end{figure*}

\textbf{Training Efficiency.}
Figure~\ref{fig:training_efficiency} compares the training dynamics of
\textbf{\emph{FutureNav}} and the policy-only baseline under the same training
schedule. \textbf{\emph{FutureNav}} consistently achieves higher SR and SPL at
matched training progress, indicating that the auxiliary world-action objectives
provide useful supervision before convergence rather than only improving the
final checkpoint. The gap is especially clear in the middle of training: at
about 42\% of the full schedule, \textbf{\emph{FutureNav}} reaches 42.3 SR and
37.8 SPL, compared with 33.2 SR and 27.9 SPL for the policy-only baseline.
Moreover, \textbf{\emph{FutureNav}} reaches a 50.2 SR after roughly 63\% of
training, already matching the final SR of the policy-only model. These results
show that world-action modeling improves both final performance and training
efficiency.

\subsection{Qualitative Analysis}

\textbf{Simulated Qualitative Analysis.}
Figure~\ref{fig:qualitative_sim} compares \textbf{\emph{FutureNav}} with the
JanusVLN~\cite{zeng2025janusvln} in simulated VLN-CE environments. Under the same instruction and
observation sequence, JanusVLN can produce reasonable local actions but is more
likely to lose spatial progress around
landmark-conditioned turns and stopping points. In contrast,
\textbf{\emph{FutureNav}} maintains a more consistent trajectory-level plan,
aligns actions with visual landmarks, and stops closer to the instructed target.
This suggests that our unified world-action modeling framework improves more than final
action classification: they help the model preserve spatial context across
multiple decision steps.

\textbf{Real-world Qualitative Analysis.}
Figure~\ref{fig:qualitative} presents zero-shot real-world navigation results of
\textbf{\emph{FutureNav}}. The model is evaluated directly in real indoor and
outdoor scenes without additional real-world navigation fine-tuning, testing
whether the simulator-trained world-action representation can transfer across
appearance, lighting, and layout shifts. The examples show that
\textbf{\emph{FutureNav}} can ground instructions to salient landmarks, maintain
the intended route over multiple actions, and make appropriate stopping
decisions near the specified destination. These results indicate that the
learned world-action modeling provides useful spatial and temporal structure for
zero-shot real-world generalization.

\section{Conclusion}

We presented \method{}, a VLM-based unified world-action modeling framework for
continuous vision-and-language navigation. Instead of training navigation only as
direct action imitation, \method{} jointly learns action prediction, inverse
dynamics, forward dynamics, and future spatial-state generation within a single
multimodal action-decoding framework. This design strengthens the model's
spatial and transition understanding during training while preserving an
efficient policy-only inference pathway. Extensive experimental results show
that \method{} achieves state-of-the-art performance with only  4B-scale
backbone, improves over prior larger VLN foundation models. Ablation studies
further confirm that the world-action objectives and spatial latent prediction
targets contribute complementary gains. Our work paves the way toward future world-action models for embodied navigation.

\bibliographystyle{unsrtnat}
\bibliography{main}


\end{document}